\PassOptionsToPackage{table}{xcolor}
\documentclass{article}
\usepackage{iclr2027_conference,times}
\usepackage[utf8]{inputenc}
\usepackage[T1]{fontenc}
\usepackage{hyperref}
\hypersetup{hidelinks}
\usepackage{url}
\usepackage{microtype}
\usepackage{graphicx}
\usepackage{subcaption}
\usepackage{booktabs}
\usepackage{amsmath,amssymb,amsthm}
\usepackage{algorithm,algorithmic}
\usepackage{float,placeins}
\usepackage{array,enumitem,multirow,makecell}
\usepackage[table]{xcolor}
\usepackage[normalem]{ulem}

\definecolor{tabgray}{gray}{0.955}
\newcommand{\effgain}[1]{{\scriptsize\textcolor{blue!70!black}{[#1]}}}
\newcommand{\speedup}[1]{\textbf{#1}}
\newcommand{\qgain}[1]{{\scriptsize\textcolor{green!45!black}{(#1)}}}
\newcommand{\qdrop}[1]{{\scriptsize\textcolor{red!70!black}{(#1)}}}
\newcommand{\qflat}[1]{{\scriptsize\textcolor{gray!70}{(#1)}}}
\newtheorem{proposition}{Proposition}

\newcommand{\mainresultstableearly}{
\begin{table}[t]
\centering
\caption{Efficiency and quality comparison of FIS-DiT and competing methods on few-step Wan 2.2 T2V and HunyuanVideo 1.5 I2V.}
\label{tab:main_results}
\footnotesize
\setlength{\tabcolsep}{2.15pt}
\renewcommand{\arraystretch}{\maintablearraystretch}
\resizebox{\textwidth}{!}{%
\begin{tabular}{lc!{\vrule width 0.35pt}ccc!{\vrule width 0.35pt}ccc}
\toprule
\multirow{2}{*}{\textbf{Method}} & \multirow{2}{*}{\textbf{Axis}} & \multicolumn{3}{c!{\vrule width 0.35pt}}{\textbf{Efficiency}} & \multicolumn{3}{c}{\textbf{VBench Evaluation}} \\
\cmidrule(lr){3-5}\cmidrule(l){6-8}
& & \makecell{\textbf{FLOPs (P)}\\$\downarrow$} & \makecell{\textbf{Latency (s)}\\$\downarrow$} & \makecell{\textbf{Speedup}\\$\uparrow$} & \makecell{\textbf{Quality Score}\\$\uparrow$} & \makecell{\textbf{Semantic / I2V Score}\\$\uparrow$} & \makecell{\textbf{Total Score}\\$\uparrow$} \\
\midrule
\multicolumn{8}{c}{\textbf{(a) Wan 2.2-T2V-A14B + Lightning LoRA} \textnormal{(81 frames, 720p, $T=4$; Quality + Semantic)}} \\
\midrule
\rowcolor{gray!10} Baseline & -- & 26.09 & 269.1 & 1.00$\times$ & 0.8846 & 0.7678 & 0.8612 \\
MagCache-Fast~\citep{magcache} & Step & 26.09 \effgain{$\downarrow0.0\%$} & 269.1 \effgain{$\downarrow0.0\%$} & 1.00$\times$ & 0.8846 \qflat{$+0.0000$} & 0.7678 \qflat{$+0.0000$} & 0.8612 \qflat{$+0.0000$} \\
MagCache-Force & Step & 19.57 \effgain{$\downarrow25.0\%$} & 203.6 \effgain{$\downarrow24.3\%$} & \speedup{1.32$\times$} & 0.8803 \qdrop{$-0.0043$} & 0.7646 \qdrop{$-0.0032$} & 0.8572 \qdrop{$-0.0040$} \\
SeaCache-Relaxed$^{\dagger}$~\citep{seacache} & Step & 19.33$^{\ddagger}$ \effgain{$\downarrow25.9\%$} & 199.3 \effgain{$\downarrow25.9\%$} & \speedup{1.35$\times$} & 0.8673 \qdrop{$-0.0173$} & 0.7468 \qdrop{$-0.0210$} & 0.8432 \qdrop{$-0.0180$} \\
TaylorSeer-Relaxed$^{\dagger}$~\citep{taylorseer} & Step & 12.92$^{\ddagger}$ \effgain{$\downarrow50.5\%$} & 133.2 \effgain{$\downarrow50.5\%$} & \speedup{2.02$\times$} & 0.8692 \qdrop{$-0.0154$} & 0.7480 \qdrop{$-0.0198$} & 0.8450 \qdrop{$-0.0162$} \\
ToMe ($r=0.3$)~\citep{tomesd} & Token & 15.17$^{\ddagger}$ \effgain{$\downarrow41.9\%$} & 156.5 \effgain{$\downarrow41.9\%$} & \speedup{1.72$\times$} & 0.8525 \qdrop{$-0.0321$} & 0.6888 \qdrop{$-0.0790$} & 0.8198 \qdrop{$-0.0414$} \\
SLA~\citep{sla} & Attn. & 11.91$^{\ddagger}$ \effgain{$\downarrow54.3\%$} & 122.9 \effgain{$\downarrow54.3\%$} & \speedup{2.19$\times$} & 0.8660 \qdrop{$-0.0186$} & 0.7115 \qdrop{$-0.0563$} & 0.8351 \qdrop{$-0.0261$} \\
\cmidrule(lr){1-8}
\textbf{FIS-Balanced} & Frame & 11.55 \effgain{$\downarrow55.7\%$} & 127.6 \effgain{$\downarrow52.6\%$} & \speedup{2.11$\times$} & \textbf{0.8845} \qflat{$-0.0001$} & 0.7610 \qdrop{$-0.0068$} & 0.8598 \qdrop{$-0.0014$} \\
\textbf{FIS-Fast} & Frame & 10.39 \effgain{$\downarrow60.2\%$} & 111.7 \effgain{$\downarrow58.5\%$} & \speedup{2.41$\times$} & 0.8837 \qflat{$-0.0009$} & 0.7546 \qdrop{$-0.0132$} & 0.8579 \qdrop{$-0.0033$} \\
\textbf{FIS-Bal. + Mag.-Force} & Hybrid & \textbf{9.87} \effgain{$\downarrow62.2\%$} & \textbf{108.7} \effgain{$\downarrow59.6\%$} & \speedup{2.48$\times$} & 0.8844 \qflat{$-0.0002$} & 0.7663 \qdrop{$-0.0015$} & 0.8608 \qflat{$-0.0004$} \\
\specialrule{0.08em}{0.4em}{0.4em}
\multicolumn{8}{c}{\textbf{(b) Wan 2.2-T2V-A14B + Lightning LoRA} \textnormal{(81 frames, 480p, $T=4$; Quality + Semantic)}} \\
\midrule
\rowcolor{gray!10} Baseline & -- & 6.71 & 66.8 & 1.00$\times$ & 0.8778 & 0.7340 & 0.8490 \\
MagCache-Fast~\citep{magcache} & Step & 6.71 \effgain{$\downarrow0.0\%$} & 66.8 \effgain{$\downarrow0.0\%$} & 1.00$\times$ & 0.8778 \qflat{$+0.0000$} & 0.7340 \qflat{$+0.0000$} & 0.8490 \qflat{$+0.0000$} \\
MagCache-Force & Step & 5.03 \effgain{$\downarrow25.0\%$} & 51.1 \effgain{$\downarrow23.5\%$} & \speedup{1.31$\times$} & 0.8732 \qdrop{$-0.0046$} & 0.7340 \qflat{$+0.0000$} & 0.8454 \qdrop{$-0.0036$} \\
SeaCache-Relaxed$^{\dagger}$~\citep{seacache} & Step & 4.76$^{\ddagger}$ \effgain{$\downarrow29.1\%$} & 47.4 \effgain{$\downarrow29.1\%$} & \speedup{1.41$\times$} & 0.8640 \qdrop{$-0.0138$} & 0.7212 \qdrop{$-0.0128$} & 0.8354 \qdrop{$-0.0136$} \\
TaylorSeer-Relaxed$^{\dagger}$~\citep{taylorseer} & Step & 3.21$^{\ddagger}$ \effgain{$\downarrow52.2\%$} & 32.0 \effgain{$\downarrow52.2\%$} & \speedup{2.09$\times$} & 0.8596 \qdrop{$-0.0182$} & 0.6935 \qdrop{$-0.0405$} & 0.8264 \qdrop{$-0.0226$} \\
ToMe ($r=0.3$)~\citep{tomesd} & Token & 4.25$^{\ddagger}$ \effgain{$\downarrow36.7\%$} & 42.3 \effgain{$\downarrow36.7\%$} & \speedup{1.58$\times$} & 0.8370 \qdrop{$-0.0408$} & 0.6672 \qdrop{$-0.0668$} & 0.8030 \qdrop{$-0.0460$} \\
SLA~\citep{sla} & Attn. & 5.05$^{\ddagger}$ \effgain{$\downarrow24.8\%$} & 50.2 \effgain{$\downarrow24.8\%$} & \speedup{1.33$\times$} & 0.8361 \qdrop{$-0.0417$} & 0.7115 \qdrop{$-0.0225$} & 0.8112 \qdrop{$-0.0378$} \\
\cmidrule(lr){1-8}
\textbf{FIS-Balanced} & Frame & 3.17 \effgain{$\downarrow52.8\%$} & 36.0 \effgain{$\downarrow46.1\%$} & \speedup{1.86$\times$} & \textbf{0.8780} \qflat{$+0.0002$} & 0.7299 \qdrop{$-0.0041$} & 0.8484 \qflat{$-0.0006$} \\
\textbf{FIS-Fast} & Frame & 2.83 \effgain{$\downarrow57.8\%$} & 32.0 \effgain{$\downarrow52.1\%$} & \speedup{2.09$\times$} & 0.8753 \qdrop{$-0.0025$} & 0.7257 \qdrop{$-0.0083$} & 0.8454 \qdrop{$-0.0036$} \\
\textbf{FIS-Bal. + Mag.-Force} & Hybrid & \textbf{2.67} \effgain{$\downarrow60.2\%$} & \textbf{30.7} \effgain{$\downarrow54.0\%$} & \speedup{2.18$\times$} & 0.8766 \qdrop{$-0.0012$} & 0.7371 \qgain{$+0.0031$} & 0.8487 \qflat{$-0.0003$} \\
\specialrule{0.08em}{0.4em}{0.4em}
\multicolumn{8}{c}{\textbf{(c) HunyuanVideo 1.5-8B + Step Distilled} \textnormal{(I2V, 81 frames, 480p, $T=8$; VBench-I2V Quality Score + I2V Score)}} \\
\midrule
\rowcolor{gray!10} Baseline & -- & 5.47 & 51.1 & 1.00$\times$ & 0.7955 & 0.9184 & 0.8570 \\
MagCache-Fast~\citep{magcache} & Step & 5.47 \effgain{$\downarrow0.0\%$} & 51.1 \effgain{$\downarrow0.0\%$} & 1.00$\times$ & 0.7955 \qflat{$+0.0000$} & 0.9184 \qflat{$+0.0000$} & 0.8570 \qflat{$+0.0000$} \\
MagCache-Force & Step & 3.42 \effgain{$\downarrow37.5\%$} & 32.6 \effgain{$\downarrow36.2\%$} & \speedup{1.57$\times$} & 0.7914 \qdrop{$-0.0041$} & 0.9012 \qdrop{$-0.0172$} & 0.8463 \qdrop{$-0.0107$} \\
SeaCache-Relaxed$^{\dagger}$~\citep{seacache} & Step & 2.72$^{\ddagger}$ \effgain{$\downarrow50.2\%$} & 25.4 \effgain{$\downarrow50.2\%$} & \speedup{2.01$\times$} & 0.7925 \qdrop{$-0.0030$} & \textbf{0.9200} \qgain{$+0.0016$} & 0.8563 \qflat{$-0.0007$} \\
TaylorSeer-Relaxed$^{\dagger}$~\citep{taylorseer} & Step & 3.30$^{\ddagger}$ \effgain{$\downarrow39.8\%$} & 30.8 \effgain{$\downarrow39.8\%$} & \speedup{1.66$\times$} & 0.7928 \qdrop{$-0.0027$} & 0.9137 \qdrop{$-0.0047$} & 0.8533 \qdrop{$-0.0037$} \\
ToMe ($r=0.3$)~\citep{tomesd} & Token & 2.29$^{\ddagger}$ \effgain{$\downarrow58.2\%$} & 21.4 \effgain{$\downarrow58.2\%$} & \speedup{2.39$\times$} & 0.6929 \qdrop{$-0.1026$} & 0.8219 \qdrop{$-0.0965$} & 0.7574 \qdrop{$-0.0996$} \\
SLA~\citep{sla} & Attn. & 3.72$^{\ddagger}$ \effgain{$\downarrow32.0\%$} & 34.8 \effgain{$\downarrow32.0\%$} & \speedup{1.47$\times$} & 0.8039 \qgain{$+0.0084$} & 0.9176 \qflat{$-0.0008$} & 0.8608 \qgain{$+0.0038$} \\
\cmidrule(lr){1-8}
\textbf{FIS-Balanced} & Frame & 2.33 \effgain{$\downarrow57.4\%$} & 24.9 \effgain{$\downarrow51.3\%$} & \speedup{2.05$\times$} & \textbf{0.8058} \qgain{$+0.0103$} & 0.9188 \qflat{$+0.0004$} & \textbf{0.8623} \qgain{$+0.0054$} \\
\textbf{FIS-Fast} & Frame & \textbf{2.06} \effgain{$\downarrow62.3\%$} & \textbf{21.3} \effgain{$\downarrow58.3\%$} & \speedup{2.40$\times$} & 0.8041 \qgain{$+0.0086$} & 0.9179 \qflat{$-0.0005$} & 0.8610 \qgain{$+0.0041$} \\
\textbf{FIS-Bal. + Mag.-Force} & Hybrid & 2.14 \effgain{$\downarrow60.9\%$} & 21.8 \effgain{$\downarrow57.3\%$} & \speedup{2.34$\times$} & 0.8049 \qgain{$+0.0094$} & 0.9183 \qflat{$-0.0001$} & 0.8616 \qgain{$+0.0047$} \\
\bottomrule
\end{tabular}%
}
\vspace{2pt}
{\scriptsize\raggedright\textit{Notes.}\ Wan reports VBench Quality, Semantic, and Total Scores ($\mathrm{Total}=0.8\,\mathrm{Quality}+0.2\,\mathrm{Semantic}$); HunyuanVideo reports VBench-I2V Quality, I2V, and Total Scores ($\mathrm{Total}=0.5\,\mathrm{Quality}+0.5\,\mathrm{I2V}$). Relative to each matched dense baseline, blue brackets denote efficiency reductions, while green/red/gray parentheses denote score improvements, degradations, or negligible changes ($|\Delta|\le0.001$). Score changes are computed before rounding. $^{\dagger}$ denotes a relaxed baseline configuration; $^{\ddagger}$ denotes idealized FLOPs estimated as $C_{\mathrm{base}}/\mathrm{Speedup}$ when traces are unavailable.\par}
\end{table}

}
\newcommand{\mainresultstablelate}{}
\newcommand{\maintablearraystretch}{0.90}
\newcommand{\ablationtablesize}{\fontsize{8pt}{9pt}\selectfont}
\newcommand{\ablationtabletabcolsep}{4.5pt}
\newcommand{\ablationtablearraystretch}{0.88}
\newcommand{\cvmathstyle}{\textstyle}
\newcommand{\cvpromptterminator}{,\qquad P=\text{number of prompts}.}
\newcommand{\cvpromptafter}{}
\newcommand{\visualcomparisonfigureearly}{}
\newcommand{\visualcomparisonfigurelate}{
\begin{figure}[H]
    \centering
    \includegraphics[width=1.00\textwidth]{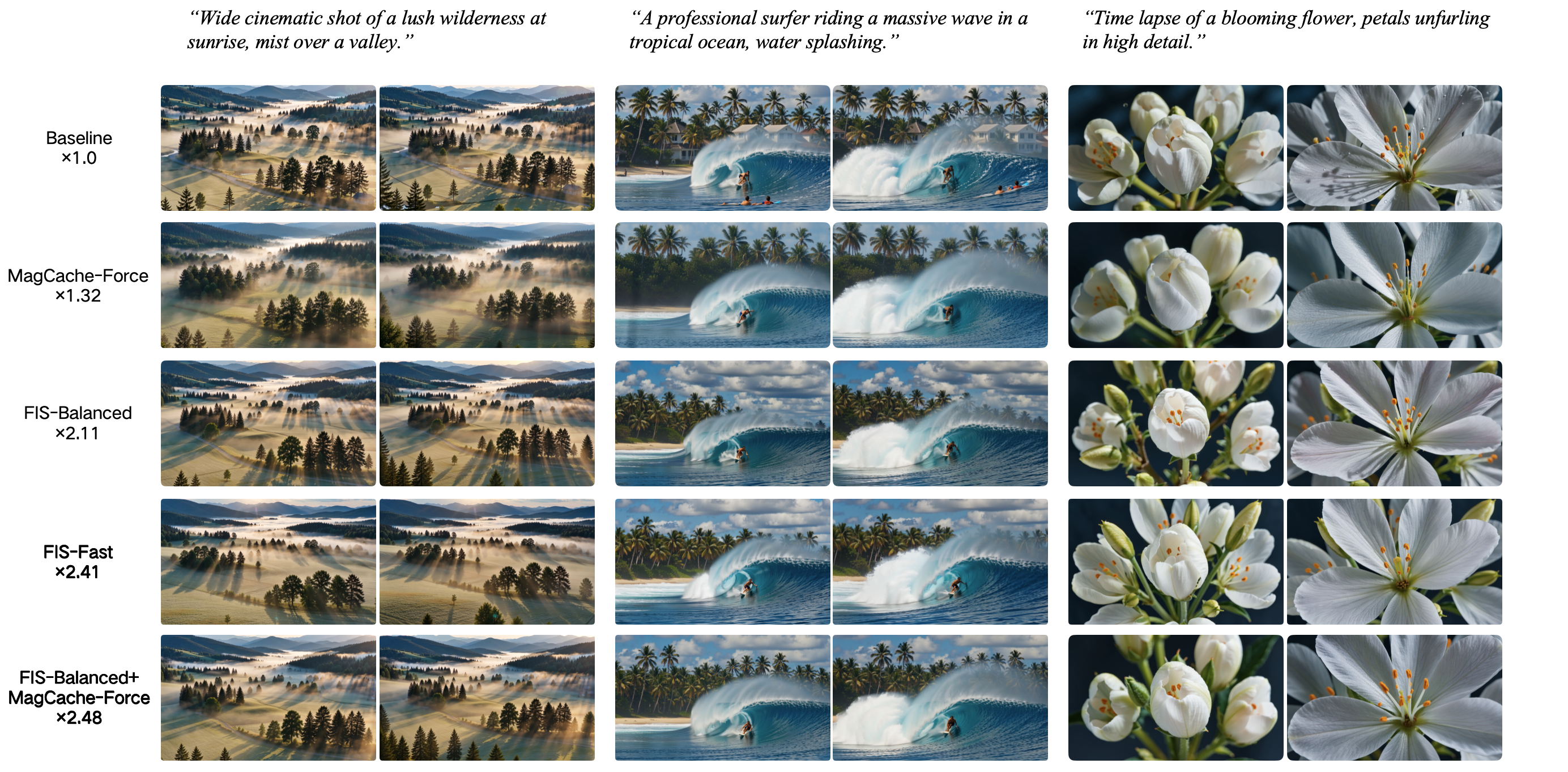}
    \caption{Qualitative comparison on 4-step Wan 2.2 at 720p across three prompts.}
    \label{fig:visual_compare}
\end{figure}

}
\newcommand{\conclusionparagraphbreak}{\par}

\newcommand{\efficiencyqualitytext}{Table~\ref{tab:main_results} shows that FIS-DiT substantially accelerates all three few-step settings with little change in aggregate VBench scores, reporting task-specific VBench Semantic and Total Scores for Wan T2V, and I2V and Total Scores for HunyuanVideo I2V. FIS-Fast reaches 2.09$\times$--2.41$\times$ speedup, with a maximum Total Score decrease of 0.0036 on Wan and no decrease on HunyuanVideo; FIS-Balanced reaches 1.86$\times$--2.11$\times$, reducing the maximum Wan decrease to 0.0014 and again showing no decrease on HunyuanVideo. Combining FIS-Balanced with MagCache-Force reaches 2.18$\times$--2.48$\times$ while keeping Total Scores close to the dense baselines, confirming complementarity with timestep caching.}

\newcommand{\visualcomparisontext}{Figure~\ref{fig:visual_compare} compares three representative prompts---misty wilderness, surfing, and blooming flower---spanning atmospheric scenes, large motion, and fine-grained details. MagCache-Force introduces reduced scene layering, structural distortion, and texture/color degradation, whereas FIS-Balanced remains close to the baseline and FIS-Fast preserves clear structures and details at over 2$\times$ speedup. FIS-Balanced + MagCache-Force also remains visually close at 2.48$\times$, consistent with the complementarity in Table~\ref{tab:main_results}.}

\newcommand{\ablationintrotext}{We ablate interleaved frame scheduling and risk-aware sparse deployment on Wan 2.2-T2V-A14B + Lightning LoRA (4-step, 480p).}

\newcommand{\riskablationtext}{Table~\ref{tab:ablation_risk_deployment} evaluates the two protection terms. Removing block-aware protection raises speedup from 1.86$\times$ to 2.04$\times$ and lowers VBench Quality Score from 0.8780 to 0.8731, consistent with the larger head/tail fluctuations in Fig.~\ref{fig:block_heterogeneity}; removing step-aware protection reaches 2.36$\times$ and 0.8586, confirming the sensitivity of final denoising steps. Removing both reaches 2.72$\times$ and 24.6s but lowers CLIP-Score to 27.35 and VBench Quality Score to 0.8427, showing that both protections are needed to balance efficiency and visual fidelity.}

\title{FIS-DiT: Breaking the Few-Step Video Inference Barrier via Training-Free Frame Interleaved Sparsity}
\author{%
  Jian Tang\textsuperscript{1}\thanks{Corresponding author.} \quad
  Jiawei Fan\textsuperscript{1} \quad
  Qingbin Liu\textsuperscript{1} \quad
  Zheng Wei\textsuperscript{1} \quad
  Jiang Bian\textsuperscript{1} \quad
  Zang Li\textsuperscript{1} \\[0.5em]
  \textsuperscript{1}Platform and Content Group, Tencent \\
  \texttt{jackyjtang@tencent.com}
}

\hypersetup{
  pdftitle={FIS-DiT: Breaking the Few-Step Video Inference Barrier via Training-Free Frame Interleaved Sparsity},
  pdfauthor={Jian Tang, Jiawei Fan, Qingbin Liu, Zheng Wei, Jiang Bian, Zang Li},
  pdfkeywords={video diffusion, DiT, sparse inference, training-free acceleration},
}

\iclrfinalcopy

\begin{document}
\maketitle
\lhead{Preprint}

\begin{abstract}
While the overall inference latency of Video Diffusion Transformers (DiTs) can be substantially reduced through model distillation, per-step inference latency remains a critical bottleneck. Existing acceleration paradigms primarily exploit redundancy across the denoising trajectory; however, we identify a limitation where these step-wise strategies encounter diminishing returns in few-step regimes. In such scenarios, fewer intermediate states can limit opportunities for feature reuse or predictive modeling, motivating complementary forms of acceleration. To overcome this, we propose Frame Interleaved Sparsity DiT (FIS-DiT), a training-free and operator-agnostic framework that shifts the optimization focus from the temporal trajectory to the latent frame dimension. Our approach is motivated by an intrinsic duality within this dimension: the existence of frame-wise sparsity that permits reduced computation, coupled with a structural consistency that calls for maintaining coverage of frame positions in the global spatiotemporal context. Leveraging this insight, we implement Frame Interleaved Sparsity (FIS) as an execution strategy that manipulates frame subsets across the model hierarchy, refreshing all latent positions without requiring full-scale block computation. Empirical evaluations on Wan 2.2 and HunyuanVideo 1.5 demonstrate that FIS-DiT reaches 2.4$\times$ speedup while maintaining comparable performance on VBench, providing a scalable and robust pathway toward real-time high-definition video generation.
\end{abstract}

\section{Introduction}

Recent advances in diffusion transformers (DiTs) have substantially improved the fidelity, motion realism, and temporal coherence of video generation~\citep{peebles2023dit,latte,opensora,cogvideox,hunyuanvideo,wan}.
Meanwhile, model distillation has greatly reduced the number of denoising steps required for high-quality synthesis, making few-step video generation increasingly practical~\citep{song2023consistency,luo2023lcm,videolcm,animatelcm,dollar}.
However, fewer sampling steps do not eliminate the inference bottleneck of DiTs.
For high-resolution video synthesis, each denoising step still requires expensive spatio-temporal Transformer computation over a large number of latent tokens.
As the denoising trajectory is compressed to only a few steps, the cost of each remaining DiT evaluation becomes an increasingly important part of overall inference latency.

A prominent family of training-free acceleration methods exploits redundancy along the denoising trajectory.
They reuse, predict, cache, or skip intermediate states across timesteps, with policies ranging from step-level reuse to block-level feature caching~\citep{deepcache,pab,fastercache,teacache,deltadit,magcache,seacache,taylorseer}.
Despite their different granularities, these methods often rely on nearby denoising states being sufficiently similar to support approximation of current-step computation.

Opportunities to exploit this redundancy may narrow in few-step distilled DiTs.
When the denoising process is compressed to only a handful of steps, fewer intermediate states are available for reuse, and adjacent steps may be more widely separated.
As a result, cross-step matching, prediction, and caching may offer less opportunity for additional acceleration in this regime.
Together, these constraints create a practical barrier to further trajectory-based acceleration in few-step settings.

To break this barrier, we shift the acceleration perspective from the denoising trajectory to the latent frame dimension.
The latent frame dimension refers to the temporal axis of the video latent representation, where each latent frame contains a complete spatial token grid.
Inspired by the strong correlation between neighboring video frames in video frame interpolation~\citep{jiang2018superslomo,bao2019dain,huang2022rife,reda2022film}, we ask whether similar redundancy also exists inside the intermediate latent representations of DiTs.
Rather than generating a low-frame-rate video followed by pixel-space interpolation, we explore whether DiTs can sparsely compute only a subset of latent frames and restore the full latent sequence within the network hierarchy.

Our analysis reveals three key properties of latent-frame dynamics in few-step DiTs.
First, adjacent latent frames within the same Transformer block exhibit structured local temporal redundancy, suggesting that omitted frame states can be approximated from neighboring ones in suitable blocks.
Second, this redundancy is block-dependent: many middle blocks are relatively stable, while some early and late blocks show larger frame-wise variations and call for protection from sparse computation.
Third, lightweight feature-level interpolation can approximate omitted latent states at low computational cost, although its error depends on the block and frame position.

Based on these observations, we propose \textbf{Frame Interleaved Sparsity DiT (FIS-DiT)}, a training-free acceleration framework for few-step DiTs. Unlike cache-based methods that reuse historical states across timesteps, FIS-DiT reduces computation within each forward pass by sparsely evaluating latent frames and reconstructing the full sequence via lightweight interpolation.

Our main contributions are summarized as follows:
\begin{itemize}[leftmargin=*, topsep=2pt, itemsep=2pt, parsep=0pt, partopsep=0pt]
\item \textbf{New Acceleration Perspective:} We observe that trajectory-based reuse may offer diminishing returns in few-step regimes and shift the focus from cross-step redundancy to within-step latent-frame sparsity.

\item \textbf{FIS-DiT:} We propose a training-free framework that sparsely computes complete latent-frame slices in stable blocks while preserving full computation in sensitive blocks, interleaves the computed frame subsets across layers to periodically refresh all frame positions, avoids additional feature-cache memory, and can be plugged into existing DiT inference pipelines without retraining.

\item \textbf{Empirical Performance:} On the challenging 4-step distilled Wan 2.2 setting at 720p, FIS-DiT achieves a 2.11$\times$--2.41$\times$ inference speedup while maintaining comparable performance on VBench.
\end{itemize}

\section{Related Work}
\label{sec:related_work}

\subsection{Diffusion Models for Video Synthesis}

Diffusion models have become foundational in generative modeling due to their ability to produce high-quality and diverse samples~\citep{sohl2015deep,ho2020ddpm,song2021score,nichol2021improved,dhariwal2021diffusion}.
Latent diffusion models improve scalability by performing denoising in a compact latent space~\citep{rombach2022ldm}, while early video diffusion models extend image diffusion backbones with temporal modules for temporally coherent generation~\citep{ho2022video,singer2022makeavideo,ho2022imagenvideo,blattmann2023videoldm,zhou2022magicvideo,guo2023animatediff}.

Recently, diffusion transformers (DiTs)~\citep{peebles2023dit} have been increasingly adopted for video generation due to their scalability and modeling capacity.
Modern video diffusion systems operate over dense spatio-temporal latent tokens and achieve strong fidelity, motion realism, and temporal coherence~\citep{wang2023lavie,videocrafter,dynamicrafter,latte,opensora,cogvideox,hunyuanvideo,wan,moviegen}.
However, this improved generation quality comes with substantial inference cost, as each denoising evaluation requires expensive Transformer computation over a large number of video latent tokens.

\subsection{Efficiency Improvements in Diffusion Models}

Existing efficiency improvements for diffusion models can be broadly categorized into reducing the number of sampling steps and reducing the computational cost per step.
For sampling step reduction, fast SDE/ODE solvers and improved samplers accelerate diffusion inference without changing model weights~\citep{karras2022edm,lu2022dpmsolver,lu2022dpmsolverpp}, while distillation and consistency-based methods compress the denoising trajectory into only a few steps~\citep{salimans2022progressive,song2023consistency,luo2023lcm}.
These ideas have also been extended to video generation~\citep{videolcm,animatelcm,dollar}.
However, few-step generation mainly shortens the denoising trajectory and does not directly reduce the cost of each remaining DiT forward pass.

Training-free acceleration methods reduce inference cost by reusing, caching, broadcasting, or predicting computation across denoising timesteps~\citep{deepcache,fasterdiffusion,pab,fastercache,teacache,deltadit,adacache,profilingdit,precisecache,foracache,magcache}. Representative designs include feature reuse or broadcasting~\citep{deepcache,pab}, DiT-oriented caching~\citep{deltadit,fastercache,teacache,adacache,profilingdit,foracache,precisecache,magcache,seacache}, and feature forecasting~\citep{taylorseer,hicachehermite}. Their reuse opportunities shrink when distillation compresses the trajectory to only a few steps. DisCa targets this regime with a learned feature-cache predictor for step-distilled video DiTs~\citep{disca}; FIS-DiT instead requires no additional training and reduces computation within each remaining step. Within-step acceleration has also been studied through token pruning, pooling, learning, and merging~\citep{rao2021dynamicvit,liang2022evit,ryoo2021tokenlearner,yin2022avit,marin2023tokenpool,bolya2022tome,diffrate}, token reduction for diffusion or multimodal inference~\citep{tomesd,chen2024fastv}, and sparse or linear attention~\citep{sla}. Fine-grained token reduction can introduce irregular indexing, alter spatial token structure, or complicate compatibility with efficient attention kernels such as FlashAttention~\citep{dao2022flashattention}. FIS-DiT is also related to video frame interpolation~\citep{jiang2018superslomo,bao2019dain,huang2022rife,reda2022film}, but reconstructs intermediate latent states inside the DiT hierarchy rather than interpolating output frames as post-processing.

\noindent\textbf{Differences with Previous Methods.}
Prior cache-based methods exploit redundancy across denoising timesteps, an assumption that becomes fragile in few-step regimes where adjacent steps contain large semantic changes. FIS-DiT shifts the acceleration axis from cross-timestep reuse to within-step latent-frame sparsity, preserving the spatial token grid and the full-frame module interface. It is also complementary to timestep-oriented caching: caching reduces redundant steps along the trajectory, while FIS-DiT reduces frame-wise computation inside each remaining step.

\section{Method}
\label{sec:method}

\subsection{Preliminaries: Video Diffusion Transformers}
\label{sec:preliminary}

\textbf{Latent Video Generation.} Modern video diffusion models operate in a compressed latent space~\citep{rombach2022ldm}. A pre-trained 3D VAE encodes a video into $z_0 \in \mathbb{R}^{F \times C \times H \times W}$, and the generative process denoises $z_T \sim \mathcal{N}(0, \mathbf{I})$ stepwise until $z_0$ is recovered.

\textbf{DiT Architecture.} The denoising network $\epsilon_\theta(z_t, t)$ is parameterized by a Transformer (DiT) with $L$ stacked blocks $\{\Phi_l\}_{l=0}^{L-1}$. Each block processes patchified tokens via sub-modules $\Phi_l^m$ (e.g., Spatial-Temporal Attention and FFNs). The global attention complexity scales quadratically with the sequence length $\mathcal{O}((F \cdot H \cdot W)^2)$, making full-frame evaluation prohibitively expensive for large $F$.

\begin{figure}[ht]
    \centering
    \begin{subfigure}[b]{0.48\linewidth}
        \centering
        \includegraphics[width=\linewidth]{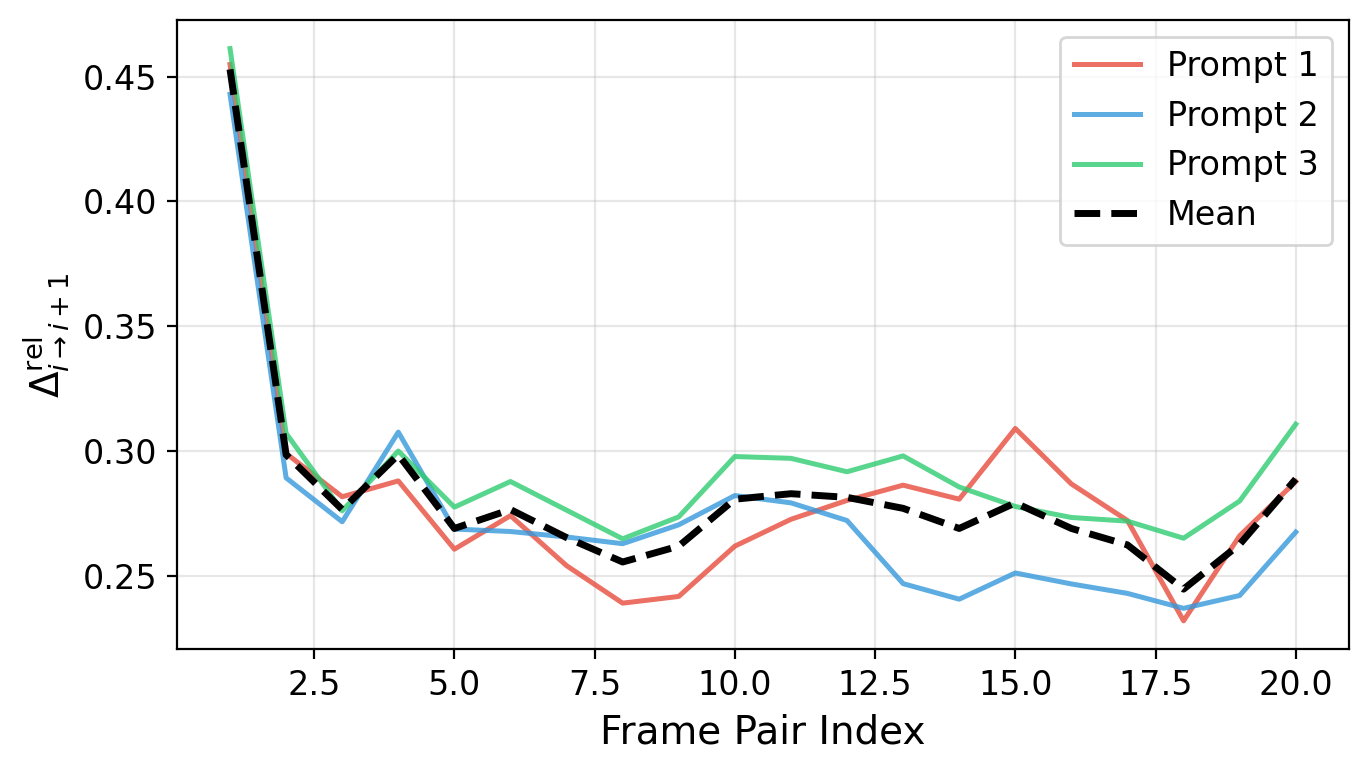}
        \caption{Wan 2.2 (40 blocks, 4 steps)}
    \end{subfigure}
    \hfill
    \begin{subfigure}[b]{0.48\linewidth}
        \centering
        \includegraphics[width=\linewidth]{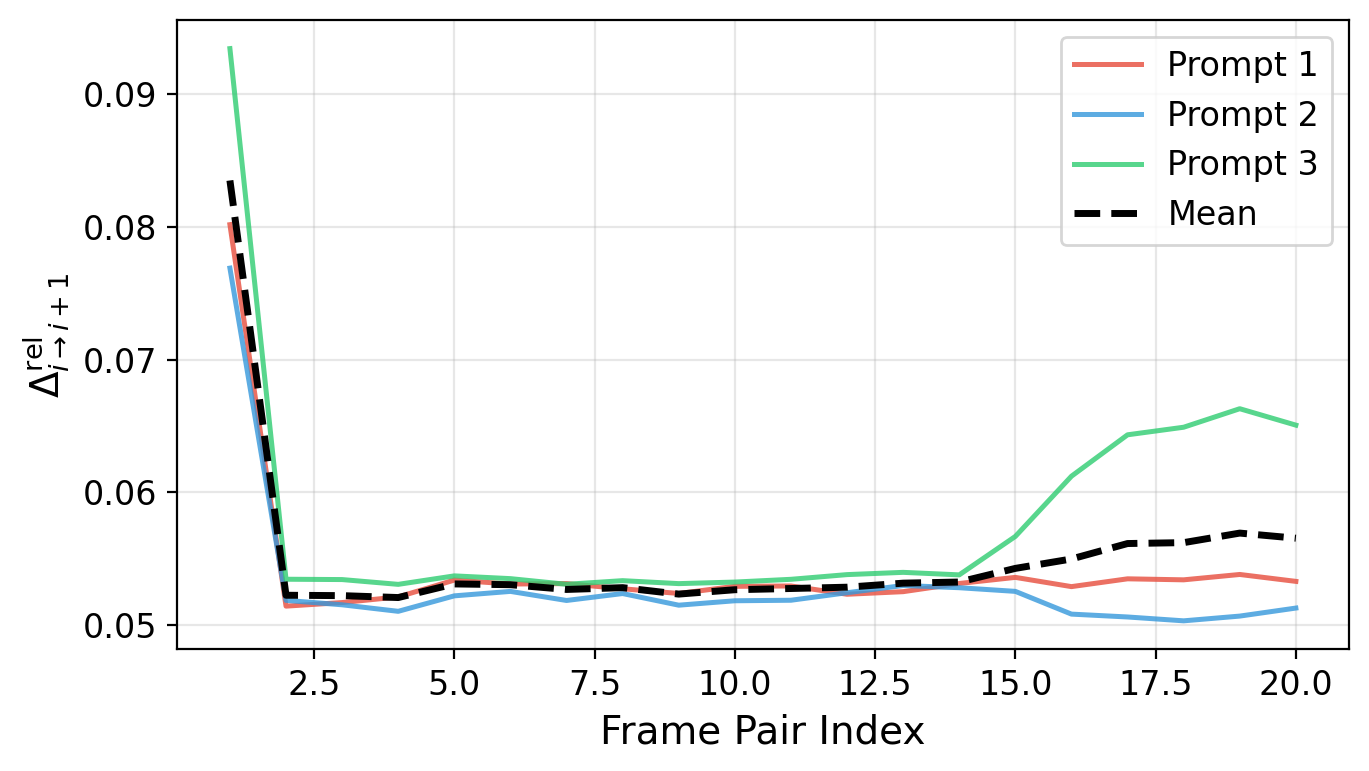}
        \caption{HunyuanVideo 1.5 (54 blocks, 8 steps)}
    \end{subfigure}
    \caption{Adjacent-frame relative changes $\Delta^{\mathrm{rel}}_{i \rightarrow i+1}$ drop after the first frame pair and then vary more mildly across time, with similar trends across three prompt types on both models.}
    \label{fig:uniform_status}
\end{figure}

\subsection{Empirical Observations: Temporal Redundancy in DiTs}
\label{sec:observations}

In this section, we empirically characterize the temporal redundancy of intermediate representations in video DiTs. In particular, we show that: \textbf{(1)} after an initial change, adjacent-frame relative-change profiles are comparatively even across time and show similar trends across three prompt types, \textbf{(2)} temporal redundancy is highly heterogeneous across network depth, and \textbf{(3)} interpolated-frame relative-error profiles show no apparent temporal drift on either backbone despite different mean levels.

Let $X_i^{(b,s,p)} \in \mathbb{R}^{H \times W \times D}$ denote the feature tensor of the $i$-th frame at transformer block $b$, denoising step $s$, and prompt $p$, where $H$ and $W$ are the spatial dimensions and $D$ is the channel dimension. For notational simplicity, we omit the superscript $(b,s,p)$ in the following definitions and assume that all quantities are computed at a fixed block, a fixed denoising step, and a fixed prompt.

To quantify frame-to-frame change, we first reduce the channel dimension by computing the $L_2$ norm of each spatial token:
\begin{equation}
    V_i(h,w) = \| X_i(h,w) \|_2,
    \label{eq:magnitude_map}
\end{equation}
where $X_i(h,w) \in \mathbb{R}^{D}$ denotes the feature vector at spatial location $(h,w)$, and $V_i \in \mathbb{R}^{H \times W}$ is the resulting spatial magnitude map of frame $i$.

We then define the \emph{absolute adjacent-frame distance} as the $L_2$ distance between two consecutive spatial magnitude maps:
\begin{equation}
    \Delta^{\mathrm{abs}}_{i \rightarrow i+1}
    =
    \|V_{i+1} - V_i\|_2.
    \label{eq:absolute_change}
\end{equation}
To obtain a scale-invariant measure, we normalize by the magnitude of the preceding frame, yielding the \emph{relative change}:
\begin{equation}
    \Delta^{\mathrm{rel}}_{i \rightarrow i+1}
    =
    \frac{\Delta^{\mathrm{abs}}_{i \rightarrow i+1}}{\|V_i\|_2+\epsilon},\qquad \epsilon=10^{-8}.
    \label{eq:relative_change}
\end{equation}
This normalized quantity measures the fractional change of the spatial feature magnitude from frame $i$ to frame $i+1$, making the comparison less sensitive to feature scale differences across blocks and denoising steps.

\noindent \textbf{Adjacent-Frame Dynamics Across Prompts.}
We evaluate Eq.~\ref{eq:relative_change} on two architecturally distinct video DiTs: Wan 2.2 (40 blocks, 4 denoising steps) and HunyuanVideo 1.5 (54 blocks, 8 denoising steps). As shown in Fig.~\ref{fig:uniform_status}, the temporal curves $\{\Delta^{\mathrm{rel}}_{i \rightarrow i+1}\}_{i=0}^{F-2}$ fall after the first frame pair and are comparatively even over the remaining pairs, so no later transition consistently dominates feature evolution. The three prompt types span fast aquatic motion, non-rigid animal motion, and slower stylized motion; their exact text is given in Appendix~\ref{app:empirical_diagnostics}. Their curves follow similar overall trends on both models, although their levels and local fluctuations differ. These measurements describe intermediate latent-feature magnitude maps after 3D VAE compression and temporal mixing, rather than pixel-level motion. This regularity supports a static evenly spaced sparse policy in the evaluated settings.

\begin{figure*}[ht]
    \centering
    \begin{subfigure}[b]{0.48\textwidth}
        \centering
        \includegraphics[width=\linewidth]{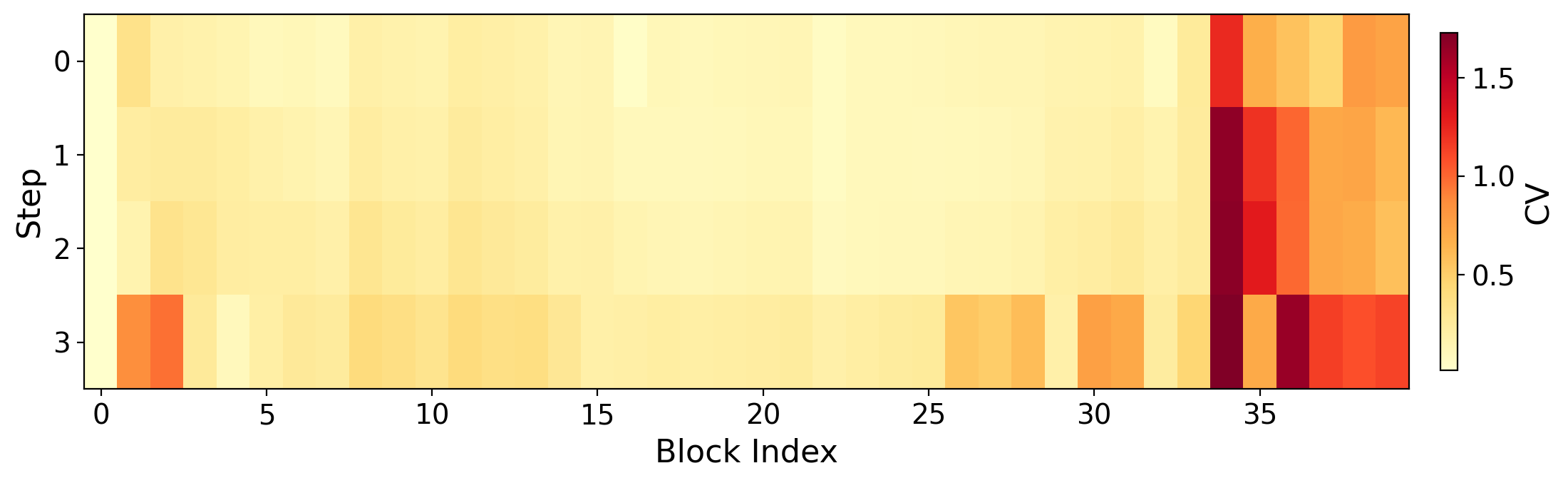}
        \caption{Wan 2.2 (40 blocks, 4 steps)}
    \end{subfigure}
    \hfill
    \begin{subfigure}[b]{0.48\textwidth}
        \centering
        \includegraphics[width=\linewidth]{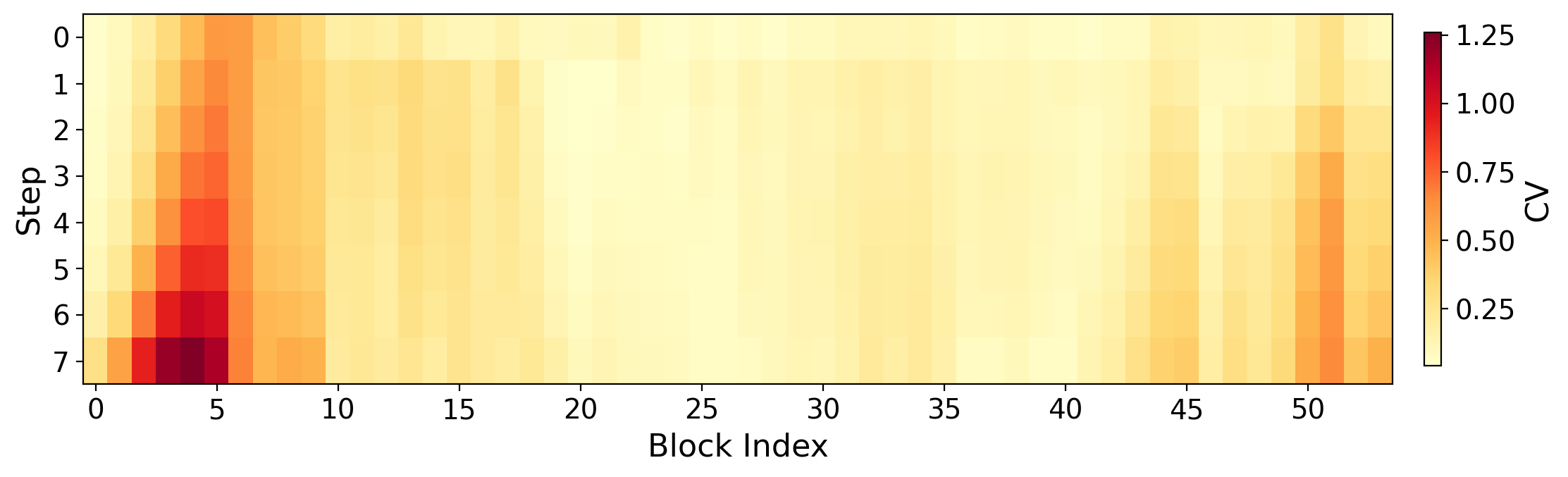}
        \caption{HunyuanVideo 1.5 (54 blocks, 8 steps)}
    \end{subfigure}
    \caption{Block-aware temporal heterogeneity. CV heatmaps show temporally stable middle blocks and larger fluctuations in head/tail blocks on (a) Wan 2.2 and (b) HunyuanVideo 1.5.}
    \label{fig:block_heterogeneity}
\end{figure*}

\noindent \textbf{Block-Aware Heterogeneity.}
Although adjacent-frame changes are broadly uniform over time, their stability varies substantially across network depth. To quantify this effect, for each block $b$ and denoising step $s$, we summarize the flatness of the temporal curve $\{\Delta^{\mathrm{rel}}_{i \rightarrow i+1}(b,s)\}_{i=0}^{F-2}$ using the coefficient of variation:
\begin{equation}
    \cvmathstyle
    \mu_{b,s}
    =
    \frac{1}{F-1}
    \sum_{i=0}^{F-2}
    \Delta^{\mathrm{rel}}_{i \rightarrow i+1}(b,s),
    \qquad
    \sigma_{b,s}
    =
    \left(
    \frac{1}{F-1}
    \sum_{i=0}^{F-2}
    \left(
    \Delta^{\mathrm{rel}}_{i \rightarrow i+1}(b,s) - \mu_{b,s}
    \right)^2
    \right)^{1/2},
\end{equation}
\begin{equation}
    \cvmathstyle
    \mathrm{CV}_{b,s}
    =
    \frac{\sigma_{b,s}}{\mu_{b,s}+\epsilon},
    \qquad
    \overline{\mathrm{CV}}_{b,s}
    =
    \frac{1}{P}
    \sum_{p=1}^{P}
    \mathrm{CV}^{(p)}_{b,s}\cvpromptterminator
    \label{eq:cv}
\end{equation}
\cvpromptafter

As shown in Fig.~\ref{fig:block_heterogeneity}, the resulting heatmaps on both models reveal a consistent block-dependent pattern: middle blocks exhibit low $\mathrm{CV}$ values, indicating highly uniform adjacent-frame changes, while head and tail blocks display noticeably larger fluctuations. This heterogeneity directly motivates our block-aware design: sparse approximation is applied only to the stable middle block set $\mathcal{M}$, while sensitive head and tail blocks retain full-frame computation.

\begin{figure}[ht]
    \centering
    \begin{subfigure}[b]{0.48\linewidth}
        \centering
        \includegraphics[width=\linewidth]{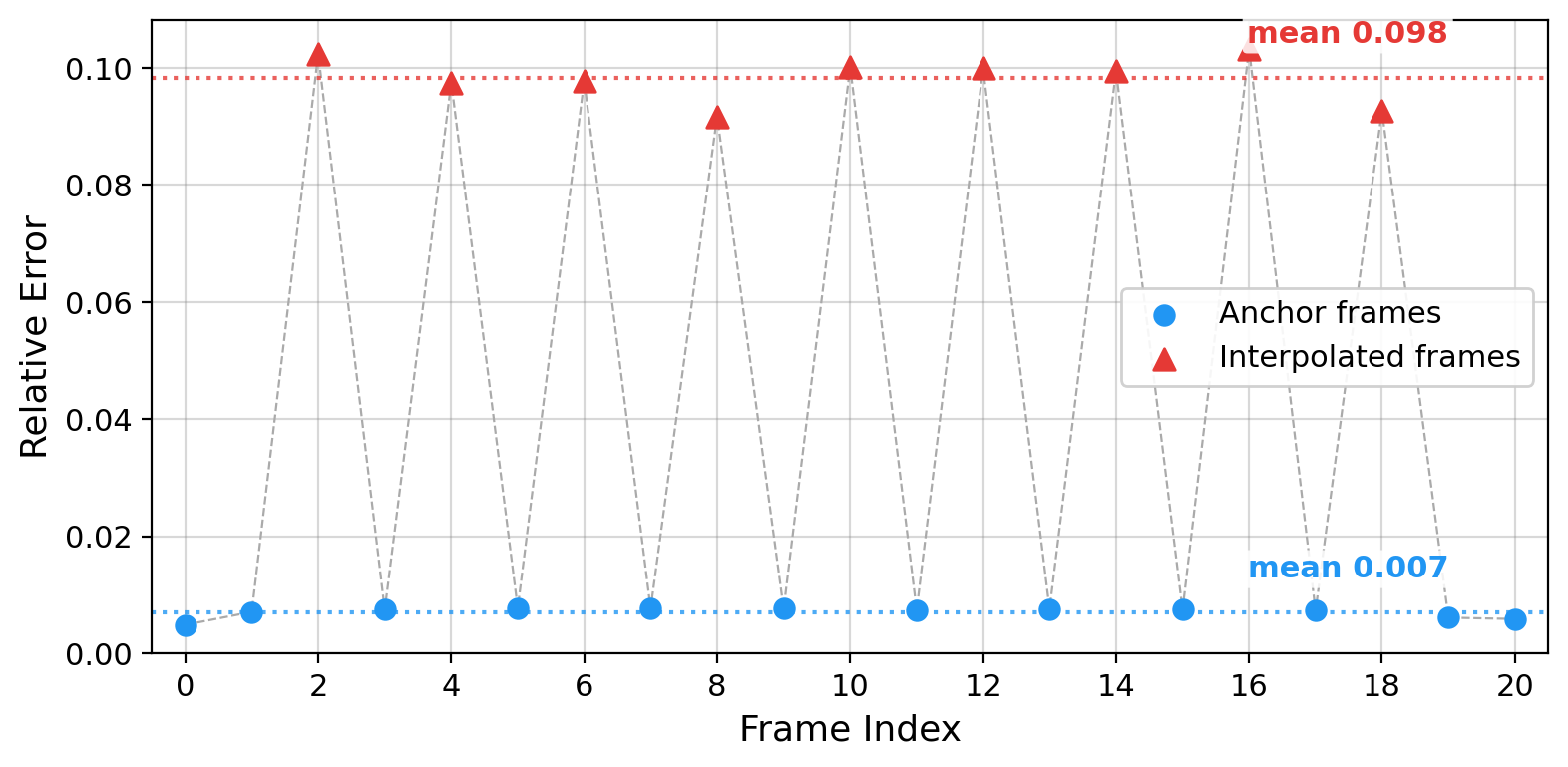}
        \caption{Wan 2.2}
    \end{subfigure}
    \hfill
    \begin{subfigure}[b]{0.48\linewidth}
        \centering
        \includegraphics[width=\linewidth]{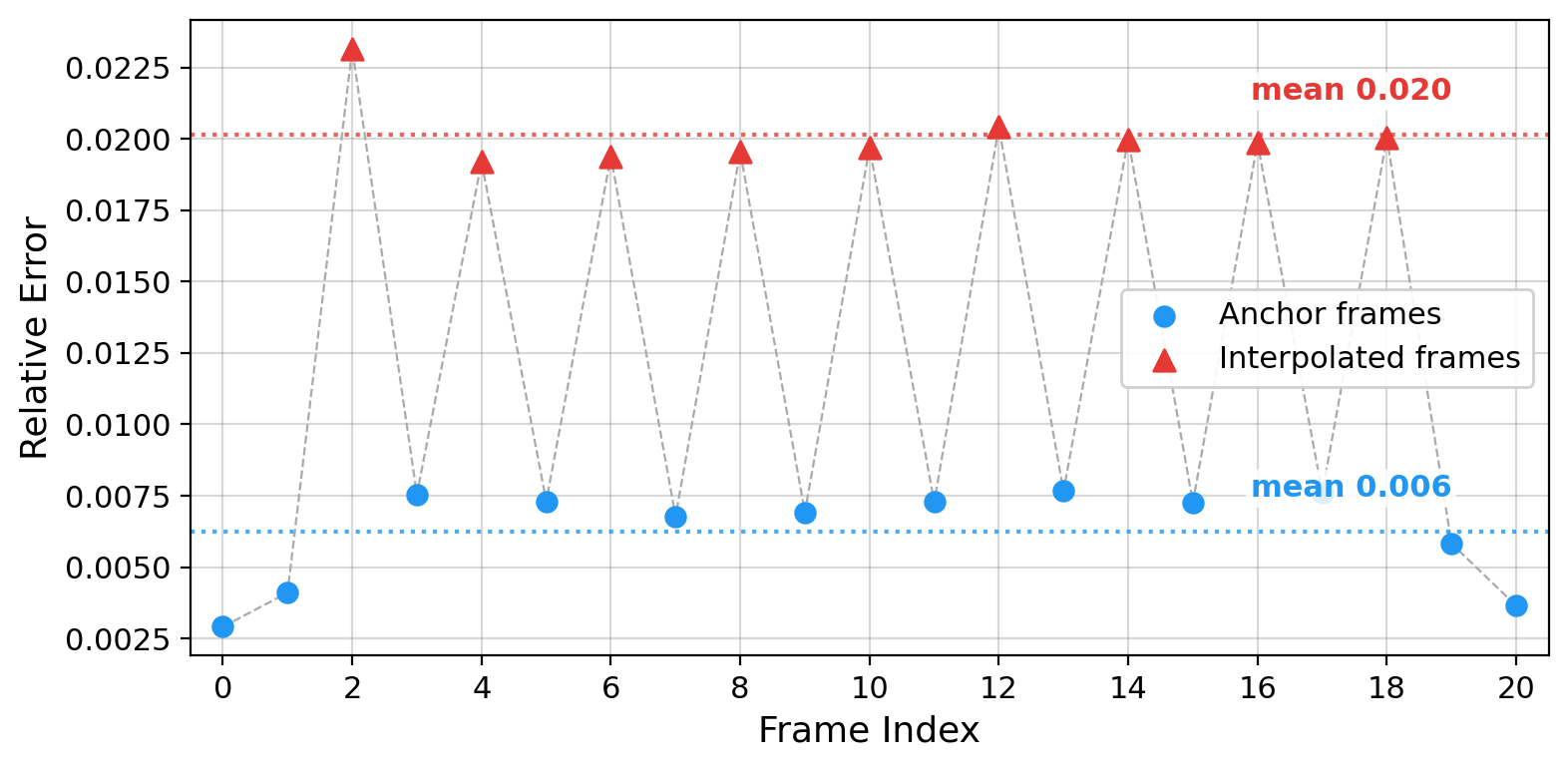}
        \caption{HunyuanVideo 1.5}
    \end{subfigure}
    \caption{Per-frame relative magnitude-map deviation. Anchor frames (blue) remain below 1\%; interpolated frames (red) show no apparent temporal drift on either model despite different mean levels.}
    \label{fig:bounded_error}
\end{figure}

\noindent \textbf{Interpolation Error Diagnostics.}
To characterize the effect of reconstructing skipped frames, we measure the per-frame relative error under sparse evaluation ($n{=}2$). Let $Y_i$ and $\hat{Y}_i$ denote the full-computation and sparse-evaluation outputs for frame $i$, respectively. The per-frame relative error is:
\begin{equation}
    \mathcal{E}_i
    =
    \frac{
    \| \hat{V}_i - V_i \|_2
    }{
    \| V_i \|_2+\epsilon
    },
    \label{eq:per_frame_error}
\end{equation}
where $V_i(h,w) = \|Y_i(h,w)\|_2$ and $\hat{V}_i(h,w) = \|\hat{Y}_i(h,w)\|_2$. As shown in Fig.~\ref{fig:bounded_error}, anchor-frame deviations remain below 1\%, while interpolated frames have mean relative deviations of 9.8\% on Wan 2.2 and 2.0\% on HunyuanVideo 1.5. Despite this difference in level, both profiles show the same alternating anchor/interpolated-frame pattern, and the interpolated-frame deviations show no apparent temporal drift. This diagnostic compares intermediate feature-magnitude maps; the reconstruction operator interpolates the full latent features, not decoded pixels. Appendix~\ref{app:empirical_diagnostics} reports the prompt-wise depth profile.

These findings show structured, block-dependent intra-step temporal redundancy with similar trends across the evaluated prompts and backbones, forming the basis of our inference strategy in Section~\ref{sec:framework}.

\begin{figure*}[t]
    \centering
    \includegraphics[width=1\textwidth]{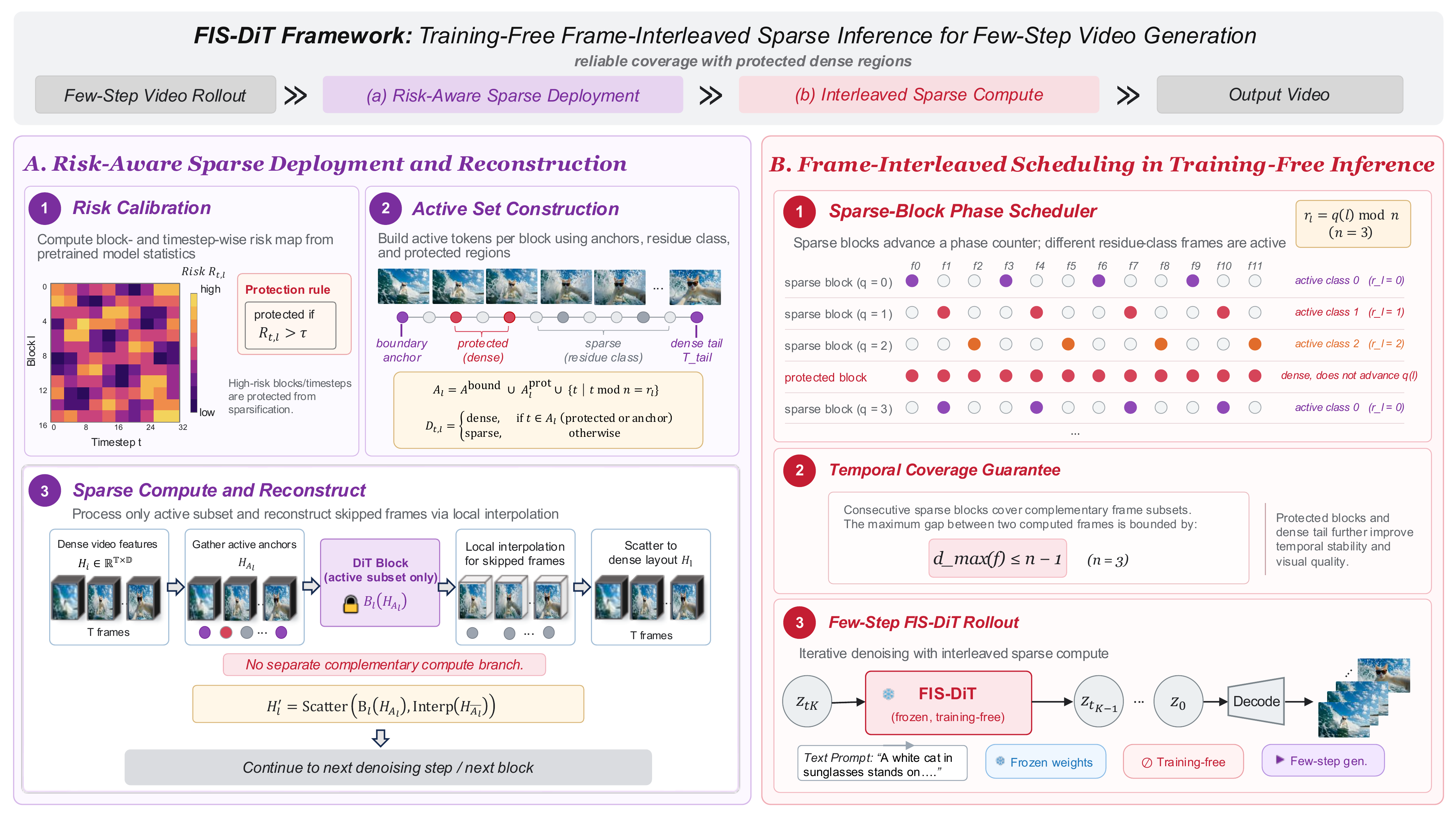}
    \caption{Overview of FIS-DiT.}
    \label{fig:framework}
\end{figure*}

\subsection{The FIS-DiT Framework}
\label{sec:framework}

FIS-DiT is a training-free sparse inference framework that replaces redundant full-frame computation with a sparse-compute-and-reconstruct paradigm. In safe regions, it computes only anchor frames, reconstructs skipped frames, interleaves anchor sets across layers, and disables sparsity in high-risk blocks or timesteps.

\subsubsection{Interpolation-based Frame Reconstruction}
\label{sec:interp_reconstruction}

Let $X_l$ be the input feature sequence of module $\Phi_l^m$ in block $l$, and let $\mathcal{F}=\{0,\dots,F-1\}$. Given anchor set $\mathcal{A}_l \subseteq \mathcal{F}$, FIS-DiT performs:
\begin{align}
    X_l^{\mathrm{anc}} &= \texttt{Gather}(X_l, \mathcal{A}_l), \label{eq:gather}\\
    Y_l^{\mathrm{anc}} &= \Phi_l^m(X_l^{\mathrm{anc}}), \label{eq:compute}\\
    X_{l+1} &= \mathcal{R}(Y_l^{\mathrm{anc}}, \mathcal{A}_l, \mathcal{F}), \label{eq:reconstruct}
\end{align}
where $\mathcal{R}$ reconstructs the full frame sequence from sparse anchor outputs.

The operator $\mathcal{R}$ is generic; since few-step video generation requires a training-free, low-overhead, and stable operator, we use local linear interpolation by default. For a skipped frame $f_u$ between neighboring anchors $f_a$ and $f_b$, we compute:
\begin{equation}
    X_{l+1}(f_u)
    =
    \frac{f_b-f_u}{f_b-f_a}Y_l^{\mathrm{anc}}(f_a)
    +
    \frac{f_u-f_a}{f_b-f_a}Y_l^{\mathrm{anc}}(f_b).
    \label{eq:linear_interp}
\end{equation}
Anchor frames are directly copied from $Y_l^{\mathrm{anc}}$. We always include the first and last frames as anchors, since boundary frames lack two-sided context and empirically show larger variations.

\paragraph{Hardware efficiency.}
Frame-level sparsity preserves contiguous spatial tokens within each selected frame, avoiding the fragmented layouts caused by unstructured token pruning. FIS-DiT can therefore reuse optimized dense kernels such as FlashAttention~\citep{dao2022flashattention} without custom CUDA kernels.

\subsubsection{Interleaved Anchor Scheduling}
\label{sec:interleaved_scheduling}

A fixed anchor set repeatedly excludes the same frames from nonlinear modules, causing error accumulation across layers. FIS-DiT instead interleaves anchor sets across sparse blocks.

Let $\kappa$ be the target anchor ratio and $n=\lceil 1/\kappa\rceil$ be the anchor stride. For sparse block $l$, let $q(l)=|\{j\in\mathcal{M}:j<l\}|$ count preceding sparse blocks, and define:
\begin{equation}
    r_l=q(l)\bmod n,
\end{equation}
The anchor set is:
\begin{equation}
    \mathcal{A}_l
    =
    \left\{
    f\in\mathcal{F}\mid (f-r_l)\bmod n=0
    \right\}
    \cup
    \{0,F-1\}.
    \label{eq:anchor_boundary}
\end{equation}

Across any $n$ consecutive sparse blocks, every non-boundary frame becomes an anchor exactly once, so no frame is reconstructed in every block of the cycle. Direct evaluation still receives features carrying earlier approximation error; it does not reset the frame to the dense trajectory.

\subsubsection{Block- and Step-Aware Sparse Deployment}
\label{sec:risk_aware_deployment}

Since approximation risk is heterogeneous, FIS-DiT applies sparse computation only when both the block and timestep are safe. We partition DiT blocks into stable middle blocks $\mathcal{M}$ and sensitive head/tail blocks $\mathcal{S}$:
\begin{equation}
    \mathcal{M}\cup\mathcal{S}=\{0,\dots,L-1\},
    \qquad
    \mathcal{M}\cap\mathcal{S}=\emptyset.
\end{equation}
Sparse reconstruction is enabled only for $l\in\mathcal{M}$, while $l\in\mathcal{S}$ retains full-frame computation.
The protected set is determined automatically by a one-time calibration pass: we compute the CV profile in Eq.~\ref{eq:cv} and apply a fixed threshold, without manual heatmap inspection or per-prompt adjustment at inference time. The resulting sets can differ across backbones under the same selection rule.

For timestep-level risk control, FIS-DiT disables sparsity in the final denoising steps, where low-noise refinement is sensitive to approximation errors. With zero-based step index $t\in\{0,\ldots,T-1\}$ and full-frame tail length $T_{\mathrm{tail}}$, the global gate is:
\begin{equation}
G(l,t)
=
\mathbf{1}[l \in \mathcal{M}]
\cdot
\mathbf{1}[t < T - T_{\mathrm{tail}}].
\label{eq:global_gate}
\end{equation}
The module computation becomes:
\begin{equation}
    X_{l+1}
    =
    \begin{cases}
    \mathcal{R}\left(
    \Phi_l^m(\texttt{Gather}(X_l,\mathcal{A}_l)),
    \mathcal{A}_l,
    \mathcal{F}
    \right), & G(l,t)=1, \\
    \Phi_l^m(X_l), & G(l,t)=0.
    \end{cases}
    \label{eq:risk_aware_compute}
\end{equation}
This gate allows FIS-DiT to accelerate stable regions while preserving full-frame computation in sensitive blocks and final denoising steps.

\section{Experiments}
\label{sec:experiments}

\subsection{Experimental Settings}
\label{sec:settings}

\paragraph{Models and Evaluation.}
We evaluate FIS-DiT on two publicly available few-step video diffusion backbones: Wan 2.2~\citep{wan} (14B, 4-step T2V at 480p/720p) and HunyuanVideo 1.5~\citep{hunyuanvideo} (8B, 8-step I2V at 480p). These models cover different scales, generation paradigms, and few-step regimes. The main comparison spans step caching (MagCache~\citep{magcache} and SeaCache~\citep{seacache}), step prediction (TaylorSeer~\citep{taylorseer}), token merging (ToMe~\citep{tomesd}), and sparse attention (SLA~\citep{sla}). We use the recent MagCache as a representative caching baseline and report its forced configuration to obtain nontrivial acceleration in the few-step setting. We likewise report relaxed SeaCache and TaylorSeer configurations. We report FLOPs, inference latency, and speedup for efficiency; for quality, we use VBench~\citep{vbench} Quality Score, Semantic Score, and Total Score for Wan T2V, and VBench-I2V~\citep{vbench,vbenchpp} Quality Score, I2V Score, and Total Score for HunyuanVideo I2V.

\mainresultstableearly

\paragraph{Implementation Details.}
All experiments are conducted on a single GPU with FlashAttention enabled: NVIDIA H20 for Wan 2.2 and NVIDIA L20 for HunyuanVideo 1.5. We evaluate 81-frame generation using 100 diverse prompts sampled from VBench~\citep{vbench}; for HunyuanVideo, each prompt is paired with a reference image for I2V evaluation.

The anchor stride $n$ controls the frame sparsity ratio, with $\kappa = 1/n$, and we evaluate $n \in \{2,3,4,5\}$. We obtain the sensitive block set $\mathcal{S}$ from a one-time CV calibration with a fixed threshold and reuse it across prompts; the resulting head/tail protection follows the pattern in Fig.~\ref{fig:block_heterogeneity}. The final denoising step is kept dense with $T_{\mathrm{tail}}=1$. Specifically, for Wan 2.2 (40 blocks, $T=4$), we set $\mathcal{S}=\{0,38,39\}$ and apply sparsity to steps 0--2. For HunyuanVideo 1.5 (54 blocks, $T=8$), we set $\mathcal{S}=\{0\text{--}5,51\text{--}53\}$ and apply sparsity to steps 0--6. We denote the conservative setting $n=3$ as \textbf{FIS-Balanced} and the aggressive setting $n=5$ as \textbf{FIS-Fast}. We also evaluate \textbf{FIS-Balanced + MagCache-Force} to test complementarity between intra-step frame sparsity and inter-step caching.

\mainresultstablelate
\visualcomparisonfigureearly

\subsection{Main Results and Visualization}
\label{sec:main_results}

\paragraph{Efficiency and Quality.}
\efficiencyqualitytext

\visualcomparisonfigurelate

\paragraph{Visual Comparison.}
\visualcomparisontext

\FloatBarrier
\subsection{Ablation Studies}
\label{sec:ablation}

\ablationintrotext

\paragraph{Effectiveness of Interleaved Frame Scheduling.}

Table~\ref{tab:ablation_interleave_stride} compares interleaved scheduling with a non-interleaved variant that uses the same computed frame subset across all sparse blocks. At matched strides and nearly identical speedups, this isolates the effect of interleaving. It raises CLIP-Score from 27.42 to 28.53 and VBench Quality Score from 0.8621 to 0.8780 at $n=3$; at $n=5$, it retains 28.50 and 0.8753, versus 26.08 and 0.8497 without interleaving, supporting its role in preventing approximation error from repeatedly concentrating on the same frames.

\begin{table}[htbp]
\centering
\caption{Ablation on interleaved frame scheduling and stride $n$.}
\label{tab:ablation_interleave_stride}
\ablationtablesize
\setlength{\tabcolsep}{\ablationtabletabcolsep}
\renewcommand{\arraystretch}{\ablationtablearraystretch}
\begin{tabular}{lccccc}
\toprule
\textbf{Frame Scheduling} & \textbf{Stride $n$} & \textbf{Latency (s)} $\downarrow$ & \textbf{Speedup} $\uparrow$ & \textbf{CLIP-Score} $\uparrow$ & \textbf{VBench Quality} $\uparrow$ \\
\midrule
\rowcolor{tabgray}
Baseline & -- & 66.8 & 1.00$\times$ & \textbf{28.56} & 0.8778 \\
w/o Interleaving & 3 & 35.7 & 1.87$\times$ & 27.42 & 0.8621 \\
w/ Interleaving & 3 & 36.0 & 1.86$\times$ & 28.53 & \textbf{0.8780} \\
w/o Interleaving & 4 & 33.4 & 2.00$\times$ & 26.61 & 0.8558 \\
w/ Interleaving & 4 & 33.8 & 1.98$\times$ & 28.51 & 0.8777 \\
w/o Interleaving & 5 & 31.6 & 2.11$\times$ & 26.08 & 0.8497 \\
w/ Interleaving & 5 & 32.0 & 2.09$\times$ & 28.50 & 0.8753 \\
\bottomrule
\end{tabular}
\end{table}

\paragraph{Impact of Risk-Aware Sparse Deployment.}
\riskablationtext

\begin{table}[htbp]
\centering
\caption{Ablation on risk-aware sparse deployment ($n=3$).}
\label{tab:ablation_risk_deployment}
\ablationtablesize
\setlength{\tabcolsep}{\ablationtabletabcolsep}
\renewcommand{\arraystretch}{\ablationtablearraystretch}
\begin{tabular}{lcccc}
\toprule
\textbf{Configuration} & \textbf{Speedup} $\uparrow$ & \textbf{Latency (s)} $\downarrow$ & \textbf{CLIP-Score} $\uparrow$ & \textbf{VBench Quality} $\uparrow$ \\
\midrule
\rowcolor{tabgray}
Baseline & 1.00$\times$ & 66.8 & \textbf{28.56} & 0.8778 \\
w/o Block-aware Protection & 2.04$\times$ & 32.8 & 28.35 & 0.8731 \\
w/o Step-aware Protection & 2.36$\times$ & 28.3 & 27.94 & 0.8586 \\
w/o Risk-aware Deployment & \textbf{2.72$\times$} & \textbf{24.6} & 27.35 & 0.8427 \\
FIS-Balanced & 1.86$\times$ & 36.0 & 28.53 & \textbf{0.8780} \\
\bottomrule
\end{tabular}
\end{table}

\section{Conclusion and Future Work}
\label{sec:conclusion}

We present FIS-DiT, a training-free framework that explores latent-frame sparsity as a complementary direction for accelerating few-step video DiTs. FIS-DiT achieves over $2\times$ inference speedup in few-step video generation while maintaining comparable VBench performance. As a sparse inference method, it naturally reduces computation and memory use without retraining, and its orthogonality to timestep-oriented acceleration enables further speedup with caching, as verified in our experiments.
\conclusionparagraphbreak
We demonstrate these results on two representative publicly available video DiT backbones, Wan 2.2 for T2V and HunyuanVideo 1.5 for I2V. Our evaluation relies on automatic metrics and does not include human-preference studies, while broader validation across video architectures remains future work. Together, these results establish latent-frame sparsity as an effective and complementary acceleration axis for few-step video generation.

\subsection*{Reproducibility statement}
The method and evaluation settings are described in Sections~\ref{sec:method} and~\ref{sec:experiments}. Appendix~\ref{app:repro_protocol} gives the comparison protocol and deployment rule, and Appendix~\ref{app:method_analysis} states the assumptions and proof for the local analysis.
\subsection*{AI use statement}
Large language models were used for language editing, formatting assistance, calculation checking, and drafting revision wording. All AI-assisted content was reviewed by the authors, who take responsibility for the final manuscript, including its technical claims, experimental results, and citations.

\clearpage
\appendix

\section{Implementation and Evaluation Details}
\label{app:repro_protocol}

\subsection{Generation and Comparison Protocol}
\label{app:model_settings}
Table~\ref{tab:model_settings_appendix} collects the three evaluated settings. Each method in a matched comparison uses the same prompt and random seed; I2V comparisons additionally use the same reference image. The main evaluation uses 100 VBench prompts. Calibration prompts are separate from evaluation prompts, and the resulting protected-block mask is reused across evaluation prompts for its backbone.

\begin{table}[htbp]
\centering
\caption{Evaluated backbones and generation settings.}
\label{tab:model_settings_appendix}
\small
\setlength{\tabcolsep}{4pt}
\begin{tabular}{lccccc}
\toprule
Backbone & Task & Frames & Resolution & Steps & GPU \\
\midrule
Wan 2.2-T2V-A14B + Lightning LoRA & T2V & 81 & 480p & 4 & H20 \\
Wan 2.2-T2V-A14B + Lightning LoRA & T2V & 81 & 720p & 4 & H20 \\
HunyuanVideo 1.5-8B + Step Distilled & I2V & 81 & 480p & 8 & L20 \\
\bottomrule
\end{tabular}
\end{table}

\paragraph{Baseline operating points.}
At their default settings, SeaCache triggers essentially no reuse at four or eight denoising steps, while TaylorSeer gives no acceleration at four steps and only weak acceleration at eight steps because of its warmup and correction-order requirements. We therefore report relaxed operating points for both methods in Table~\ref{tab:baseline_configs}; neither method's original paper targets few-step generation. The corresponding rows in Table~\ref{tab:main_results} are marked with $\dagger$.

\begin{table}[htbp]
\centering
\caption{Operating-point summary for the evaluated methods.}
\label{tab:baseline_configs}
\small
\setlength{\tabcolsep}{3.6pt}
\begin{tabular}{lp{0.27\linewidth}p{0.43\linewidth}}
\toprule
Method & Acceleration axis & Configuration used in this paper \\
\midrule
MagCache-Fast & Step caching & Default setting \\
MagCache-Force & Step caching & Force-cache the eligible denoising step that best satisfies MagCache's threshold criterion, ensuring at least one cached step \\
SeaCache-Relaxed & Step caching & Wan: \texttt{threshold=10, warmup=1, end\_step=3, verbose=True, boundary=0.0}; HunyuanVideo: \texttt{threshold=1.0, warmup=1, end\_step=7} \\
TaylorSeer-Relaxed & Feature forecasting & Wan: \texttt{order=0, warmup=1, compute\_interval=3, boundary=0.9}; HunyuanVideo: \texttt{warmup=2} (default order 1, interval 2) \\
ToMe & Token merging & Merge ratio $r=0.3$ \\
SLA & Sparse attention & Matched few-step setting \\
FIS-Balanced & Frame sparsity & Stride $n=3$, calibrated protected mask, dense final step \\
FIS-Fast & Frame sparsity & Stride $n=5$, same protected mask and dense final step \\
\bottomrule
\end{tabular}
\end{table}

\paragraph{Efficiency accounting.}
Table~\ref{tab:main_results} reports measured end-to-end latency separately from FLOPs. On Wan 480p, FIS-Balanced reduces FLOPs by 52.8\% and latency by 46.1\%; the difference reflects work outside the sparsified modules and other execution overhead. One-time CV calibration is a setup cost and is not included in the per-video latency.

\subsection{One-Time Calibration and Deployment}
\label{app:calibration_algorithm}
The calibration rule is applied once to calibration prompts; it does not update model weights or require prompt-specific statistics at inference time. Let $\overline{\mathrm{CV}}_{b,s}$ be the prompt-averaged CV from Eq.~\ref{eq:cv}. To retain both block-wise and step-wise risk information while obtaining structured masks, we project this two-dimensional profile onto the block and denoising-step axes,
\begin{equation}
r_b=\frac{1}{T}\sum_{s=0}^{T-1}\overline{\mathrm{CV}}_{b,s},
\qquad
u_s=\frac{1}{L}\sum_{b=0}^{L-1}\overline{\mathrm{CV}}_{b,s},
\qquad
\widehat r_b=\frac{r_b}{\max_j r_j},
\qquad
\widehat u_s=\frac{u_s}{\max_v u_v}.
\label{eq:block_risk}
\end{equation}
The normalization makes both summaries insensitive to the absolute CV scale of a backbone. To encode the observed head--middle--tail structure along network depth, let $\delta_b=\min\{b,L-1-b\}/(L-1)$ be the normalized distance to the nearest network boundary and define the boundary-aware block score
\begin{equation}
B_b=\frac{\widehat r_b}{\delta_b}\quad (1\le b\le L-2),
\qquad B_0=B_{L-1}=+\infty.
\label{eq:boundary_cv_score}
\end{equation}
An interior block is eligible when $B_b\le\tau_B$. To avoid isolated sparse islands, we select the longest contiguous eligible block interval,
\begin{equation}
(a,c)=\underset{1\le i\le j\le L-2}{\arg\max}\;(j-i+1)
\quad\text{s.t.}\quad B_b\le\tau_B\ \ \forall b\in\{i,\ldots,j\},
\label{eq:structured_block_mask}
\end{equation}
and apply the same structural constraint along the denoising trajectory by selecting the longest contiguous eligible step interval,
\begin{equation}
(p,q)=\underset{0\le i\le j\le T-1}{\arg\max}\;(j-i+1)
\quad\text{s.t.}\quad \widehat u_s\le\tau_T\ \ \forall s\in\{i,\ldots,j\}.
\label{eq:structured_step_mask}
\end{equation}
For either axis, ties are resolved first by the lower mean score and then by the smaller starting index. We set $\mathcal{M}_B=\{a,\ldots,c\}$, $\mathcal{M}_T=\{p,\ldots,q\}$, and use their Cartesian product as the structured sparse region. For visualizing the joint block--step risk before this structural projection, we define
\begin{equation}
R_{b,s}=\max\left\{\frac{B_b}{\tau_B},\frac{\widehat u_s}{\tau_T}\right\},
\qquad
G(b,s)=\mathbf{1}[b\in\mathcal{M}_B]\,\mathbf{1}[s\in\mathcal{M}_T],
\label{eq:joint_cv_gate}
\end{equation}
so locations with $R_{b,s}>1$ are protected, while the interval constraints prevent fragmented sparse masks.

Around the selected block boundaries, Wan has $B_1=9.795$, $B_{37}=9.363$, and $B_{38}=20.267$; HunyuanVideo has $B_5=10.600$, $B_6=6.613$, $B_{50}=8.125$, and $B_{51}=16.033$. Consequently, both block masks are unchanged for any common threshold $\tau_B\in[9.795,10.600)$; we use the round default $\tau_B=10$. Along the step axis, Wan has $\max_{s\le2}\widehat u_s=0.640$ and $\widehat u_3=1$, while HunyuanVideo has $\max_{s\le6}\widehat u_s=0.909$ and $\widehat u_7=1$. We use $\tau_T=0.95$, which lies in the common stable interval between the largest eligible-step score and the final-step score. The resulting intervals are $\mathcal{M}_B=\{1,\ldots,37\}$ and $\mathcal{M}_T=\{0,1,2\}$ for Wan 2.2, and $\mathcal{M}_B=\{6,\ldots,50\}$ and $\mathcal{M}_T=\{0,\ldots,6\}$ for HunyuanVideo 1.5. Thus, both axes remain contiguous, and the selected step intervals correspond to $T_{\mathrm{tail}}=1$ in Section~\ref{sec:settings}.

\begin{algorithm}[htbp]
\caption{CV calibration and FIS-DiT deployment.}
\label{alg:calibration_deployment}
\small
\begin{algorithmic}[1]
\STATE \textbf{Input:} calibration prompts $\mathcal{D}_{\rm cal}$, stride $n$, fixed thresholds $\tau_B=10$ and $\tau_T=0.95$
\STATE Run one dense statistics pass on $\mathcal{D}_{\rm cal}$ and calculate $B_b$ and $\widehat u_s$ using Eqs.~\ref{eq:block_risk}--\ref{eq:boundary_cv_score}.
\STATE Find $(a,c)$ and $(p,q)$ using Eqs.~\ref{eq:structured_block_mask}--\ref{eq:structured_step_mask}; freeze $\mathcal{M}_B=\{a,\ldots,c\}$ and $\mathcal{M}_T=\{p,\ldots,q\}$.
\FOR{each denoising step $t$ and block $l$}
    \IF{$l\notin\mathcal{M}_B$ or $t\notin\mathcal{M}_T$}
        \STATE Evaluate the block densely.
    \ELSE
        \STATE Set $r_l=q(l)\bmod n$; gather and evaluate anchors $\mathcal{A}_l$; reconstruct skipped-frame outputs.
    \ENDIF
\ENDFOR
\end{algorithmic}
\end{algorithm}

\section{Additional Method Analysis}
\label{app:method_analysis}

\subsection{Local Interpolation Error}
\label{app:interp_error_bound}
Let $X_b(\tau)$ be a continuous interpolation of the dense feature trajectory over frame coordinate $\tau$ within block $b$. For neighboring anchors $a<c$, suppose $X_b$ is twice differentiable on $[a,c]$ and $\sup_{\tau\in[a,c]}\|X_b''(\tau)\|_2\le M_b$. Linear interpolation of the \emph{dense} anchors yields, for $a<f<c$,
\begin{equation}
\|X_b(f)-\widetilde X_b(f)\|_2
\le \frac{M_b}{2}(f-a)(c-f)
\le \frac{M_b}{8}(c-a)^2.
\label{eq:interp_error_bound}
\end{equation}
FIS-DiT computes anchors under a sparse frame context. If their deviations from dense anchors are $\epsilon_a$ and $\epsilon_c$, then with $\lambda=(f-a)/(c-a)$ its reconstructed output satisfies
\begin{equation}
\|X_b(f)-\widehat X_b(f)\|_2
\le \frac{M_b}{8}(c-a)^2+(1-\lambda)\epsilon_a+\lambda\epsilon_c.
\label{eq:interp_error_with_context}
\end{equation}
Thus the interpolation component grows quadratically with anchor spacing under this smoothness assumption, while sparse-context error is a separate term. These are local feature bounds; neither equation bounds final-video quality for the nonlinear denoising network.

\paragraph{What CV measures.}
CV in Eq.~\ref{eq:cv} measures how uneven adjacent-frame relative changes are within a block and step. A low CV means those changes have similar magnitudes; it does not imply small reconstruction error or provide an upper bound on $M_b$. The protected-block rule uses CV as an empirical indicator of temporal non-uniformity. This distinction also explains why the depth-wise relative errors in Table~\ref{tab:depth_error_appendix} can vary across blocks despite broadly low CV in middle blocks.

\subsection{Interleaving and Reconstruction Depth}
\label{app:error_propagation}
Let $e_l(f)$ be the deviation of frame $f$ from the dense trajectory entering block $l$. A first-order local expansion gives
\begin{equation}
e_{l+1}(f)\approx J_l(f)e_l(f)+\delta_l^{\rm ctx}(f)+\delta_l^{\rm int}(f),
\label{eq:error_recurrence}
\end{equation}
where $J_l$ is the local Jacobian, $\delta_l^{\rm ctx}$ is sparse-context perturbation, and $\delta_l^{\rm int}$ is the new interpolation residual. Selecting $f$ as an anchor removes $\delta_l^{\rm int}(f)$ for that block, though prior and context errors may remain.

\begin{proposition}[Bounded consecutive reconstruction depth]
Under the phase rule $r_l=q(l)\bmod n$ and anchor set in Eq.~\ref{eq:anchor_boundary}, each non-boundary frame is directly evaluated once in every $n$ consecutive sparse blocks. It is therefore reconstructed in at most $n-1$ consecutive sparse blocks.
\label{prop:reconstruction_depth}
\end{proposition}
\begin{proof}
The sparse-block counter $q(l)$ increases by one at each sparse block. Across $n$ consecutive sparse blocks, $q(l)\bmod n$ visits every residue in $\{0,\ldots,n-1\}$ exactly once. Any frame $f$ matches one residue $f\bmod n$ and is an anchor for that phase. The other at most $n-1$ phases reconstruct it. Boundary frames are anchors in every phase.
\end{proof}
Indexing the phase by absolute block number would lose this property when protected blocks interrupt the sparse sequence. Proposition~\ref{prop:reconstruction_depth} limits consecutive \emph{new interpolation} at each frame; it does not reset existing error or imply a global bound without assumptions on the Jacobian norms in Eq.~\ref{eq:error_recurrence}.

\subsection{Approximate Compute Accounting}
\label{app:compute_model}
An idealized sparse-block cost can be written as
\begin{equation}
c(n)\approx (1-\alpha-\beta)+\frac{\alpha}{n}+\frac{\beta}{n^2},
\end{equation}
where $\alpha$ and $\beta$ represent cost components approximately linear and quadratic in the selected frame fraction, respectively. If a fraction $\rho_B$ of blocks and $\rho_T$ of denoising steps are eligible, the approximate normalized DiT cost is $1-\rho_B\rho_T(1-c(n))+C_{\rm interp}$, where $C_{\rm interp}$ accounts for gather, scatter, and reconstruction. This model describes why protected blocks and dense tail steps limit the gain from a larger stride. Measured latency also includes hardware and pipeline overhead, as reflected in Table~\ref{tab:main_results}.

\section{Empirical Diagnostics: Prompt Set and Depth-wise Error}
\label{app:empirical_diagnostics}

Figure~\ref{fig:uniform_status} uses three prompt types: fast aquatic motion (P0), non-rigid animal motion (P1), and slower motion with a stylized background (P2). The exact prompts are:
\begin{itemize}
    \item P0: ``A shark swimming in clear Caribbean ocean'';
    \item P1: ``a cat playing in park'';
    \item P2: ``A boat sailing leisurely along the Seine River with the Eiffel Tower in background by Vincent van Gogh''.
\end{itemize}
Table~\ref{tab:depth_error_appendix} reports the prompt-wise relative deviation of intermediate feature-magnitude maps on Wan 2.2, averaged over sparse denoising steps. The error rises in the middle blocks and returns toward the entry level near the exit for all three prompts. CV describes the uniformity of adjacent-frame changes, so a low CV need not imply a low reconstruction deviation. These observed depth profiles describe intermediate approximation behavior; they do not establish a global bound on final-video error.

\begin{table}[ht]
\centering
\caption{Depth-wise relative feature-magnitude-map deviation on Wan 2.2, averaged over sparse denoising steps.}
\label{tab:depth_error_appendix}
\small
\setlength{\tabcolsep}{5pt}
\begin{tabular}{lccc}
\toprule
Prompt & Entry (2--9) & Middle (18--25) & Exit (32--39) \\
\midrule
P0 -- shark & 4.1\% & 12.6\% & 5.3\% \\
P1 -- cat & 4.3\% & 12.2\% & 4.4\% \\
P2 -- painterly boat & 4.0\% & 12.4\% & 3.7\% \\
\midrule
\textbf{Mean} & \textbf{4.1\%} & \textbf{12.4\%} & \textbf{4.4\%} \\
\bottomrule
\end{tabular}
\end{table}

\paragraph{Qualitative examples.}
Figure~\ref{fig:visual_compare} uses the following exact prompts: ``Wide cinematic shot of a lush wilderness at sunrise, mist over a valley''; ``A professional surfer riding a massive wave in a tropical ocean, water splashing''; and ``Time lapse of a blooming flower, petals unfurling in high detail.'' They represent an atmospheric scene, large motion, and continuous fine-grained change, respectively.

\FloatBarrier

\end{document}